\documentclass[10.7pt, ]{article}
\usepackage[letterpaper, margin=2.54cm, top=2.54cm]{geometry}
\usepackage{lmodern}
\usepackage{authblk} 
\usepackage{amssymb,amsmath}
\usepackage{wrapfig}
\usepackage{graphicx,grffile}
\usepackage[labelfont=bf,labelsep=period]{caption}
\usepackage{ifxetex,ifluatex}
\usepackage{fixltx2e} 
\usepackage{float}
\usepackage{changepage}
\usepackage[dvipsnames]{xcolor}
\usepackage{setspace}
\usepackage{algorithm,algpseudocode}

\usepackage{array}
\newcolumntype{C}[1]{>{\centering\let\newline\\\arraybackslash\hspace{0pt}}m{#1}}
\newcolumntype{x}[1]{>{\centering\arraybackslash\hspace{0pt}}p{#1}}

\ifnum 0\ifxetex 1\fi\ifluatex 1\fi=0 
  \usepackage[T1]{fontenc}
  \usepackage[utf8]{inputenc}
\else 
  \ifxetex
    \usepackage{mathspec}
  \else
    \usepackage{fontspec}
  \fi
  \defaultfontfeatures{Ligatures=TeX,Scale=MatchLowercase}
\fi
\IfFileExists{upquote.sty}{\usepackage{upquote}}{}
\IfFileExists{microtype.sty}{%
	\usepackage{microtype}
	\UseMicrotypeSet[protrusion]{basicmath} 
}{}

\usepackage{lipsum} 

\usepackage{sectsty}
\allsectionsfont{\fontsize{11}{11}\selectfont\uppercase}
\subsectionfont{\bfseries\fontsize{11}{11}\selectfont}
\subsubsectionfont{\fontsize{11}{11}\selectfont\normalfont}

\makeatletter 
\renewcommand\@biblabel[1]{#1.} 
\makeatother %

\usepackage{etoolbox}
\makeatletter
\patchcmd{\@maketitle}{\LARGE}{\bfseries\fontsize{15}{16}\selectfont}{}{}
\makeatother

\makeatletter
\def\maxwidth{\ifdim\Gin@nat@width>\linewidth\linewidth\else\Gin@nat@width\fi}
\def\maxheight{\ifdim\Gin@nat@height>\textheight\textheight\else\Gin@nat@height\fi}
\makeatother

\setkeys{Gin}{width=\maxwidth,height=\maxheight,keepaspectratio}
\ifx\paragraph\undefined\else
\let\oldparagraph\paragraph
\renewcommand{\paragraph}[1]{\oldparagraph{#1}\mbox{}}
\fi
\ifx\subparagraph\undefined\else
\let\oldsubparagraph\subparagraph
\renewcommand{\subparagraph}[1]{\oldsubparagraph{#1}\mbox{}}
\fi

\usepackage{hyperref}
\hypersetup{
	unicode=true,
	pdftitle={My Cool Title Here},
	pdfauthor={Author One, Author Two, Author Three},
	pdfkeywords={keyword1, keyword2},
	pdfborder={0 0 0},
	breaklinks=true
}
\title{\vspace{-2em} A Method to Automate the Discharge Summary Hospital Course for Neurology Patients}
\author[ ]{\bf\fontsize{12}{13}\selectfont Vince C. Hartman, MS\textsuperscript{1}, Sanika S. Bapat, MS\textsuperscript{1}, Mark G. Weiner, MD\textsuperscript{2,3}, Babak B. Navi, MD\textsuperscript{4}, Evan T. Sholle, MS\textsuperscript{3} Thomas R. Campion, Jr., PhD\textsuperscript{3,5}}
\affil[1]{\fontsize{12}{13}\selectfont Cornell Tech, New York, New York; Emails: vch6@cornell.edu, sb2644@cornell.edu}
\affil[2]{\fontsize{12}{13}\selectfont Department of Medicine, Weill Cornell Medicine, New York, New York}
\affil[3]{\fontsize{12}{13}\selectfont Department of Population Health, Weill Cornell Medicine, New York, New York}
\affil[4]{\fontsize{12}{13}\selectfont Department of Neurology and Feil Family Brain and Mind Research Institute, Weill Cornell Medicine, New York, New York}
\affil[5]{\fontsize{12}{13}\selectfont Clinical \& Translational Science Center, Weill Cornell Medicine, New York, New York}
\date{} 

\begin{document}
\maketitle

\section{Abstract}\label{abstract}
\emph{Generation of automated clinical notes have been posited as a strategy to mitigate physician burnout. In particular, an automated narrative summary of a patient’s hospital stay could supplement the hospital course section of the discharge summary that inpatient physicians document in electronic health record (EHR) systems. In the current study, we developed and evaluated an automated method for summarizing the hospital course section using encoder-decoder sequence-to-sequence transformer models. We fine-tuned BERT and BART models and optimized for factuality through constraining beam search, which we trained and tested using EHR data from patients admitted to the neurology unit of an academic medical center. The approach demonstrated good ROUGE scores with an R-2 of 13.76. In a blind evaluation, two board-certified physicians rated 62\% of the automated summaries as meeting the standard of care, which suggests the method may be useful clinically. To our knowledge, this study is among the first to demonstrate an automated method for generating a discharge summary hospital course that approaches a quality level of what a physician would write.}

\section{Objective}
\label{sec:objective}
Physicians spend approximately two hours in the electronic health record (EHR) for every one hour of patient care \cite{arndt2017tethered}. The time required for recording, reviewing and summarizing information in EHRs has imposed complex and burdensome workflows on physicians, which has contributed to burnout \cite{downing2018physician, shanafelt2015changes}. To alleviate documentation burden \cite{quiroz2019challenges, martin2019using}, multiple efforts have pursued automated summary of the hospital patient record through natural language processing (NLP) \cite{hunter2008summarising, shing2021towards, alsentzer2018extractive}.

When a patient is discharged from a hospital, a physician authors a discharge summary, a transition of care document that summarizes the patient’s hospital stay and is sent to providers who continue the patient’s care in other settings \cite{kripalani2007deficits}. While the discharge summary is both required and valuable, clinical workflow can delay its availability \cite{roughead2011continuity}, which can increase the risk of rehospitalization \cite{van2002effect} and medication errors \cite{bergkvist2009improved}. In the United States, the content of this document can only include information that has already been documented within the EHR \cite{kind2008documentation}. In generating a discharge summary, physicians spend most of their time manually writing the hospital course section, a textual narrative that describes the progress of treatment for the patient from admission to discharge. Automating this section could potentially save time for physicians, as note templates have automated other sections of the discharge summary \cite{dean2016design} but not the hospital course section \cite{adams2021s}. 

In a prior study, we demonstrated the feasibility of automating the hospital course section of a discharge summary \cite{hartman2022discharge}. However, the study employed the MIMIC-III dataset \cite{johnson2016mimic}, which was limited to intensive care unit (ICU) patients and did not cover the full hospital stay. Furthermore, the prior study only measured textual overlap between the automated and reference summaries but not quality nor factuality. In the current study, we developed a novel method for generating clinical summaries using EHR data from inpatient neurology hospitalizations. We evaluated performance with state-of-the-art benchmarks and physician experts.

\section{Background and Significance}
\label{sec:works}
Text summarization can be categorized into two subdomains, extraction and abstraction. Extraction identifies key terms and phrases and concatenates them to form a summary, whereas abstraction generates new sentences to synthesize a summary. Generally, abstraction is more fluent and coherent. Until about 2017, clinical text summarization was mainly through extraction \cite{syed2021survey}; abstraction required substantial time and domain expertise with unpromising results \cite{hunter2008summarising}. Abstractive text summarization has accelerated due to applications of deep learning models called transformers, especially Bidirectional Encoder Representations from Transformers (BERT) \cite{devlin2018bert} and Bidirectional and Auto-Regressive Transformers (BART) \cite{lewis2019bart}.

BERT has a bidirectional language representation structure that overcomes restrictions with unidirectional language representation models \cite{devlin2018bert, vaswani2017attention}. With BERT, the entire input sequence is fed in at once unlike previous deep learning models, such as recurrent neural networks (RNNs), which read text sequentially.  

BART was created specifically for abstractive text summarization \cite{lewis2019bart}. BART uses a BERT autoencoder with noisy masked input data so as to force a decoder to denoise and reconstruct the original text. Encoder-decoder sequence-to-sequence models, such as BART, are very effective for sequence generation tasks such as text summarization \cite{rothe2020leveraging}. Currently, BART is the best performing open source model for text summarization on the CNN/Daily Mail and XSum data sets \cite{rohde2021hierarchical}.  

In the domain of clinical summarization, Yalunin et al. presented a Longformer encoder and BERT decoder for an abstractive summary of patient hospitalization histories \cite{yalunin2022abstractive}. Additional studies used physician notes rather than the entirety of structured and unstructured data available in the EHR: Shing et al. demonstrated that automation of discharge summaries was possible but noted issues of factuality \cite{shing2021towards}, and Cai et al. showed an approach for automating a patient-facing After-Visit Summary when provided a clinical note summary \cite{cai2022generation}.   In a related effort, Gao et al. demonstrated that encoder-decoder sequence-to-sequence transformers can summarize a patient’s primary diagnostic problems from a current progress note \cite{gao2022summarizing}. Despite these studies demonstrating summarization of physician notes, physicians may desire a summary note generated from their prior notes and not from their current note that is being authored.

Krishna et al. demonstrated the ability to generate Subjective, Objective, Assessment and Plan (SOAP) notes through an extractive-abstractive summarization pipeline \cite{krishna-etal-2021-generating}. In a similar paper, Joshi et al. showed that a pointer generator network with a penalty can be used to summarize medical conversations with 80\% of relevant information captured \cite{joshi-etal-2020-dr}.

While nearly all EHR summarization studies have used automated measures such as ROUGE scores to evaluate performance, Zhang et al. engaged physician experts to measure clinical validity of summaries \cite{zhang2019optimizing}. Because automated metrics do not address grammar and consistency, physician evaluations using a Likert-scale can improve understanding of automated summary quality, readability, factuality, and completeness. 

This study employs techniques from the previous works: BERT and BART models as well as both automated and physician scoring. We expand on the prior research by being the first to present a unique method to automate the discharge summary hospital course.  

\section{Materials \& Methods}
\label{sec:methods}

\subsection{Data collection}
\label{sec:dataset}

From an institutional repository containing data from EHR and other source systems \cite{campion2022architecture}, we obtained a dataset consisting of 6,600 hospital admissions from 5,000 unique patients admitted to the inpatient neurology unit at NewYork-Presbyterian/Weill Cornell Medical Center, a 2,600-bed quaternary-care teaching hospital in New York City affiliated with Weill Cornell Medicine of Cornell University. We focused on neurology patients because the speciality is known to exhibit higher clinical complexity as compared to a general inpatient patient; neurology patients have  20\% more interactions with physicians, 27\% more comorbidities, spend 81\% more days in the hospital, and have a 16\% higher mortality rate than the general inpatient patient \cite{tonelli2018complexity}. All patients had a hospitalization with a length of stay of at least 48 hours between the years of 2010 and 2020.  The dataset contained at least one admit note and one discharge summary per patient, which is a regulatory requirement for each hospital admission \cite{kind2008documentation, jointcommission_hp}. Each record of the dataset contained a combination of demographics and clinical details as described in Table \ref{tab:clinical-data-types}.

\begin{table}[t]
\small
\centering
\caption{\label{tab:clinical-data-types}Data types used from the dataset\\}
\begin{tabular}{|p{30mm}|p{98mm}|}
\hline
\textbf{Type of Data} & \textbf{Description} \\ \hline
Descriptive & Age, sex, marital status, race, mortality status \\ \hline
Encounters & Admission date, admission diagnoses (ICD-10 code and description), discharge date, discharge disposition \\ \hline
Free text documents & Admission notes, emergency department provider notes, progress notes, consult notes, operative reports, pathology reports, radiology reports, discharge summaries \\ \hline
Measurements & Laboratory results (LOINC), vital signs \\ \hline
\end{tabular}
\end{table}

For each discharge summary, we extracted the hospital course section using a regular expression. The corpus of hospital course sections served as the gold standard for model development and evaluation. We created train, validation, and test data sets with a ratio of 80:10:10. 

Furthermore, we extracted physician follow-up sentences from the hospital course sections using a BERT model trained on the CLIP dataset \cite{mullenbach2021clip}. The 64,748 follow-up sentences were then reviewed and annotated with a label of 1 or 0 signifying if they were follow-ups for future plans of care. For all the free text documents in Table \ref{tab:clinical-data-types} with the exclusion of discharge summaries, we also created a separate annotated dataset of 71,115 documents with a label of 1 or 0 signifying if the document had any content captured that was considered to be salient or not for creating the discharge summary.

\subsection{Abstractive pipeline summarization approach}
\label{sec:abstractive-approach}

When using transformers with smaller datasets, the recommended best approach is to fine-tune a pre-trained model \cite{devlin2018bert}. Pre-trained transformers generally have a maximum input token length of 512 or 1024, which is challenging to summarize the full patient record \cite{sun2019fine}. To overcome this limitation, three different strategies have traditionally been used: (1) truncating the document by, for example, taking only the first 512 tokens as input \cite{zhang2020bertal}, (2) employing a neural network that scales sequentially, such as a recurrent neural network (RNN) with a transformer output or a transformer with attention that scales with sequence length \cite{beltagy2020longformer}, or (3) summarizing or extracting individual sections first and then performing another layer of summarization during the merging of those sections \cite{gehrmann2018bottom, gidiotis2020divide}. The preferred strategy in the medical domain is to summarize smaller individual sections and combine those sections with a second layer of summarization \cite{shing2021towards, adams2021s}. This approach is preferred since it more closely resembles the physicians’ current workflow where they extract salient information first and then synthesize it into a narrative summary \cite{feblowitz2011summarization}. Similar to this ensemble strategy, we employed a “day-to-day approach” \cite{hartman2022discharge} that overcomes the limitations of long-form documents in transformers by summarizing individual clinical notes per day and concatenating them to form a clinical narrative summary.

\begin{figure}[t]
    \centering
    \caption{Data flow that shows how EHR data is segmented into three separate sections through the following transformer models referred to as the “day-to-day approach”: (1) HPI summarization, (2) daily narrative document classification and summarization, and (3) follow-up extraction classification. The automated summary  is constructed by chronologically assembling the results.}
    \includegraphics{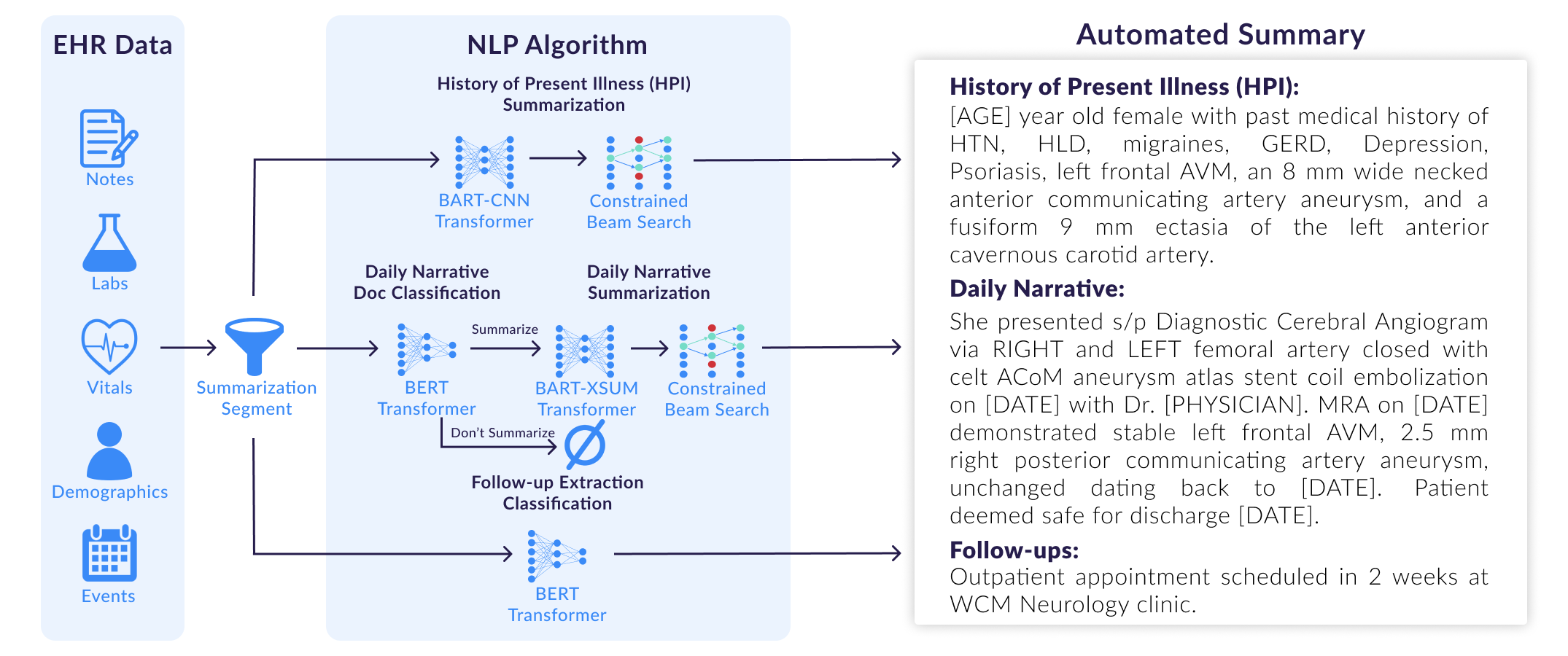}
    \label{fig:data_flow}
\end{figure}

As illustrated in Figure \ref{fig:data_flow}, we segmented the initial dataset into three parts: (1) history of present illness (HPI) summarization which is primarily constructed from the admission note, (2) daily narrative document classification and summarization which chronologically details the patient’s full course of treatment, and (3) follow-up extraction classification which identifies any follow-ups to occur at a subsequent outpatient encounter that are documented in any clinical notes within 72 hours of discharge
 (Table \ref{tab:day-approach-examples}). Each of these three parts ingests only specific types of clinical notes as a means to limit the total amount of input words into the transformer models.

\begin{table}[t!]
\small
\centering
\caption{\label{tab:day-approach-examples}Example sentences from the summarization dataset for the three separate parts of the day-to-day approach. Protected health information has been redacted.\\}
\begin{tabular}{|p{28mm}|p{100mm}|}
\hline
\textbf{Segment} & \textbf{Example Sentence} \\ \hline
History of Present Illness (HPI) & Pt is a [AGE] year old woman with atrial fibrillation (on lovenox), HTN, HLD, hypothyroidism, s/p sternotomy (uncertain cardiothoracic history), who presented as a transfer for management of Covid pneumonia. \\ \hline
Daily Narrative & Subsequently, overnight 2/5, she became hypertensive to systolic 170s and mildly tachycardic; fentanyl was titrated up to 300mcg/h given concern for inadequate sedation while paralyzed on roc gtt at 6.5. \\ \hline
Follow-ups & She will follow up with Dr. [Physician] as an outpatient. \\ \hline
\end{tabular}
\end{table}

\subsection{Constrained beam search for medical summarization}
\label{sec:constrain}

An ongoing research concern of abstractive summarization models is that they can hallucinate text that is not consistent with the source documents which creates factuality problems \cite{cao2018faithful}. Abstractive summarization models are trained at the word-level to minimize cross entropy compared with a reference summary, but this is not functionally equivalent to maximizing factuality. Researchers in the field of NLP have investigated some of the following approaches to improve factuality: measuring the inconsistencies through a question answering metric \cite{Durmus_2020}, ranking summary correctness through textual entailment predictions \cite{falke-etal-2019-ranking}, using graph-based attention \cite{zhu2020enhancing}, rewarding factuality through reinforcement-learning \cite{zhang2019optimizing}, and constraining beam search during model inference \cite{king2022don}. We implemented the  latter approach, constrained beam search, as a means to improve factuality in our models. By constraining beam search, inconsistent medical terms are reduced which is the most concerning hallucination to correct for a clinical summary as seen in the example in Table \ref{tab:constrain}.
\begin{table}[t!]
\small
\caption{\label{tab:constrain}
A motivational example of how clinical summaries can hallucinate. Inconsistent medical terms are highlighted in red font. In this example, the proposed BART model with constrained beam search for medical terminology removes the clinical inconsistencies from the HPI summary. \\}
\centering
\begin{tabular}{p{32mm}|p{96mm}}
\hline
 \textbf{Source} & \textbf{Text} \\ \hline
Admission Note &  55-year-old male with history of two vessel CAD, ischemic cardiomyopathy, EF 15\%, mitral regurgitation, and diabetes on oral agents who presents from OSH s/p VT/VF cardiac arrest... \\ \hline
BART & 55-year-old male with history of two vessel CAD, ischemic cardiomyopathy, EF 15\%, \textcolor{red}{altered mental status} and \textcolor{red}{hypotension}, now s/p VT/VF cardiac arrest... \\ \hline
BART constrained & 55-year-old male with history of two vessel CAD, ischemic cardiomyopathy, EF 15\%, \textcolor{ForestGreen}{mitral regurgitation}, and \textcolor{ForestGreen}{diabetes} who presents from OSH  s/p VT/VF cardiac arrest... \\
\hline
\end{tabular}
\end{table}

At inference, sequence to sequence transformer models are generally paired with the heuristic algorithm of beam search \cite{meister2020best}. Beam search allows for multiple candidate summaries for comparison based on a beam width and conditional probabilities. The best beam is then chosen based on a logarithmic score. Our approach constrains traditional beam search by penalizing the logarithmic score of any words from a set of banned words (Algorithm \ref{alg:beamsearch}). For constructing our set of banned words, we first used a custom medical dictionary $V_M$ derived from SNOMED CT \cite{bodenreider2004unified}. For any medical words and synonyms that intersect both the source document $x$ and $V_M$, we permit them during beam search; otherwise all other words in the medical dictionary are banned in the generated sequence (the set of banned words) $W_{banned}$ and the next best alternative word is selected as the output. Similar approaches have been found to provide an improvement in the factuality of the generated summary with only a slight decrease in its fluency \cite{king2022don}.
\begin{algorithm}[t!]
\footnotesize
\caption{Constrained Beam Search Approach}
\label{alg:beamsearch}
\begin{algorithmic}
\State $V_M \gets \mathrm{medical \ vocabulary}$
\State $x \gets \mathrm{clinical \ note \ (source \ document)}$
\State $x_M \gets V_M \cap x $
\State $x_M \gets  x_M \ + \ \text{GetSynonyms}(x_M)$
\State $W_{banned} \gets V_M - x_M $
\State
\State $y := \{ \ y_{seq} \ , \ y_{score} \ \}$
\For{$\beta$   \textbf{in} beam width}
    \While { $y_{\beta_{seq}}^{-1} \neq $ \textless end\textgreater}
    \State $y_{\beta_{seq}}^{(i)} \ , \ y_{\beta_{score}}^{(i)} := \text{BeamSearch}(x, y_{\beta_{seq}})$
    \If {$y_{\beta_{seq}}^{(i)} \in \ W_{banned}$ }
    \State \Return {$ y_{\beta_{score}} := -\infty$}
    \EndIf
    \EndWhile
\EndFor
\State \Return {$y_{\beta_{seq}}$ such that $\max (y_{score})$}
\end{algorithmic}
\end{algorithm}

\subsection{Models}
\label{sec:models}
Our approach is coupled with two state-of-the-art NLP models, BERT and BART. We use BERT for classification and BART for text summarization. Additionally, in our evaluation, we used TextRank algorithm \cite{mihalcea2004textrank} as a baseline method for comparison with the transformers.

We fine-tuned two BERT models on two separate datasets - free text documents the model should or should not summarize and the follow-up sentences - with a maximum input length of 512 tokens respectively for 3 epochs.

We used the pre-trained BART-CNN and BART-XSum transformer models from HuggingFace. We fine-tuned the models on the summarization dataset for 3 epochs with a maximum input token length of 1024.

\subsection{Evaluation}
\label{sec:evaluation}

We measured the performance of the summarization tasks (HPI summarization, daily narrative summarization, and discharge summary hospital course summarization) and the classification tasks (daily narrative document classification, follow-up extraction classification)  as seen in Figure \ref{fig:data_flow}. Additionally, we measured physician perception of quality, readability, factuality, and completeness.

\subsubsection{Summarization tasks}
\label{sec:summarization-tasks-eval}

For each of the summarization tasks, we compared BART, BART with beam search constrainment, and TextRank models through ROUGE scores, word count, and error rates.  Of note, for the final task of summarizing the discharge summary hospital course, we compared TextRank, the day-to-day approach without beam search constrainment, and the day-to-day approach with beam search constrainment (Figure \ref{fig:data_flow}).

ROUGE recall scores measure the textual overlap between the automated and physician-written summary \cite{shing2021towards, adams2021s, zhang2019optimizing, lewis2019bart}. We reported ROUGE scores on a scale from 0 to 100 where a higher score indicates better summarization performance. For longform document summarization tasks such as our study, state-of-the-art ROUGE recall scores are within the ranges of 39-51 for ROUGE-1 (R-1), 10-24 for ROUGE-2 (R-2), and 36-46 for ROUGE-L (R-L) \cite{rohde2021hierarchical, xiong2022adapting}.

We measured conciseness through the average word count and standard deviation (sd.) of the automated summary in comparison to the physician-written summary \cite{choudhry2016readability, myers2006discharge}. Conciseness was important since if the summaries were too long, they could lose relevance to downstream outpatient providers.

To understand the effectiveness of constraining medical terminology during beam search for improving factuality, we used a dependency arc entailment (DAE) model that was pre-trained on XSum \cite{goyal2021annotating}. We generated sentence level summaries for both the HPI and daily narrative summarization tasks and DAE was used to calculate both word and sentence level error rates (the fraction that was determined to be non-factual). Of note, since hospital charts have many words, we were not able to calculate word and sentence level error rates for discharge summary hospital course summarization due to limits with the DAE model.

\subsubsection{Classification tasks}
\label{sec:classification-tasks-eval}

We measured accuracy, recall, precision, and F1-scores for the document classification and follow-up extraction models. Statistics were captured from a binary label of 0 and 1 for both models. The labels measured either (1) correctly identifying the document that should or should not be summarized; and (2) correctly selecting the follow-up sentences that should or should not be included in the hospital course summary.

\subsubsection{Physician perception}
\label{sec:perception-tasks-eval}
To measure quality, readability, factuality, and completeness of an automated summary \cite{zhang2019optimizing, bernal2018impact}, two board-certified physicians (MGW, an internist; and BBN, a neurologist) blindly rated 25 pairs of patient discharge summaries, one generated by the automated method and one written by a physician for a particular hospitalization. The summaries were randomly selected from the test dataset and were ordered randomly so neither physician could ascertain whether they were reviewing the computer- or physician-generated summary. Each physician rated all 50 summaries on a Likert scale of 1 to 10 where 1 was poor and 10 was excellent with the criteria of each metric listed in Table \ref{tab:assessment-criteria} in the Appendix. Note that our quality measure effectively encompassed a summary being simultaneously concise, readable, factual, and complete.  Additionally, both physicians agreed that a Likert score of 7 for quality indicated that the summary met overall sufficient clinical validity. 

To measure inter-rater reliability of the scores between the two physicians, we used intraclass correlation coefficient (ICC) with a two-way random effects model and consistency, which measured for the degree of similarity among the two reviewers and their ratings; it has a range of 0 to 1, where 1 represents unanimous agreement and 0 indicates no agreement \cite{gisev2013interrater}.

\section{Results}
\label{sec:results}

\subsection{Summarization tasks}
\label{sec:summarization-tasks-results}
As shown in Table  \ref{tab:rouge}, the BART-based approaches had higher ROUGE scores indicative of better performance compared to the TextRank baseline in all three sub-tasks. Given that the HPI task for TextRank had a high ROUGE-2 of 28.94 as a baseline, the implication was that there is high textual overlap between the source notes and produced summary for the HPI segment. Higher textual overlap has been shown in other studies to assist with maintaining factuality \cite{ladhak2021faithful}. Although BART with constrainment had slightly lower ROUGE scores than BART without constrainment, BART with constrainment had lower word and sentence error rates.  Stated differently, we observed a tradeoff between ROUGE scores and error rates, which is consistent with the literature \cite{king2022don}. The overall performance of the day-to-day-approach with constrainment (Figure \ref{fig:data_flow}) had a ROGUE-2 score of 13.76, which was within the lower range of other state-of-the-art longform document summarization models \cite{xiong2022adapting}.

\begin{table}[t!]
\centering
\caption{\label{tab:rouge}ROUGE recall scores \textit{\{R-1/R-2/R-L\}}, Word count (WC) mean and standard deviation (Sd.), and word and sentence error rates (ER) as seen with a dependency arc entailment (DAE) model are presented  for our proposed models for comparing automated and physician-written summaries. Sentences within each summarization section are individually compared (HPI and daily) along with the full hospital course summary. Note - hospital charts had 81.5k words on average, which is too large for the DAE model.}
\begin{tabular}{p{50mm}x{10mm}x{10mm}x{10mm}x{22mm}C{17mm}C{16mm}}
\hline
\textbf{Summarization Task} & \textbf{R-1} & \textbf{R-2} & \textbf{R-L} & \textbf{\scriptsize Word Count Mean (+- Sd.)} & \textbf{\scriptsize Word-ER $\downarrow$} & \textbf{\scriptsize Sent-ER $\downarrow$} \\ \hline
 \hline
 \multicolumn{7}{l}{\textbf{History of Present Illness
(HPI)}}
\\\hline \hline
Baseline: Textrank & 43.30 & 28.94 & 38.40 & 53 (+-9) & 0.3 & 0.7 \\
BART & 61.67 & 53.12 & 59.69 & 43 (+-9) & 6.3 & 42.4 \\
BART constrained & 61.02 & 52.78 & 59.05 & 44 (+-11) & 5.7 & 40.6\\ \hline
\hline
\multicolumn{7}{l}{\textbf{Daily Narrative}}
\\\hline \hline
Baseline: Textrank & 9.55 & 1.32 & 8.93 & 20 (+-3) & 31.0 & 42.2 \\
BART & 46.59 & 35.03 & 43.95 & 10 (+-7) & 9.9 & 28.8 \\ 
BART constrained & 46.42 & 34.77 & 43.75 & 11 (+-13) & 7.2 & 26.7\\ \hline
\hline
\multicolumn{7}{l}{\textbf{Discharge Summary Hospital Course}}
\\\hline \hline
Baseline: Textrank & 15.48 & 4.18 & 8.51 & 511 (+- 451) &---&--- \\
Day-to-day & 37.10 & 14.44 & 19.64 & 444 (+-374) &---&--- \\
Day-to-day constrained & 35.97 & 13.76 & 18.83 & 421 (+-365) &---&--- \\
\hline
\end{tabular}
\end{table}

The physician-written reference summaries were highly condensed with 591 words on average (Sd. +/- 597 words) per hospital chart with 81.5k words on average (Sd. of +/- 150.3k words). Conciseness results of our  model can be seen in Table 4 with the average word count of 421 words for the discharge summary hospital courses; thus, our day-to-day approach had shorter summaries in length than physician-written ones by 170 words on average.

\subsection{Classification tasks}
\label{sec:classification-tasks-results}
For the two models for the day-to-day approach for classification, results for accuracy, recall, precision, and F1-scores are in Table \ref{tab:accuracy}. The daily narrative document classification task had an accuracy of 78.83\% at determining which clinical documents should or should not be included for summarization for the discharge summary hospital course. And the follow-up sentences task had an accuracy of 96.11\%, which implied the model was very effective at identifying which follow-up sentences should or should not be included in the hospital course section.

\begin{table}[t!]
\small
\centering
\caption{\label{tab:accuracy} Accuracy, recall, precision, and F1-score of the classification models of the day-to-day approach.\\}
\begin{tabular}{p{44mm}x{19mm}x{16mm}x{18mm}x{16mm}}
\hline
 \textbf{Classification Task} & \textbf{Accuracy} & \textbf{Recall} & \textbf{Precision} & \textbf{F1} \\ \hline
Daily Narrative Document & 0.7883 & 0.4390 & 0.7500 & 0.5538 \\
Follow-up Sentences & 0.9611 & 0.7851 & 0.8364 & 0.8100 \\
\end{tabular}
\end{table}

\subsection{Physician perception}
\label{sec:physician-perception-results}
\begin{figure}[t!]
    \centering
    \caption{Box plot for the Likert scores for quality, readability, factuality, and completeness of the automated and physician-written summaries by the physician reviewers.\\}
    \includegraphics{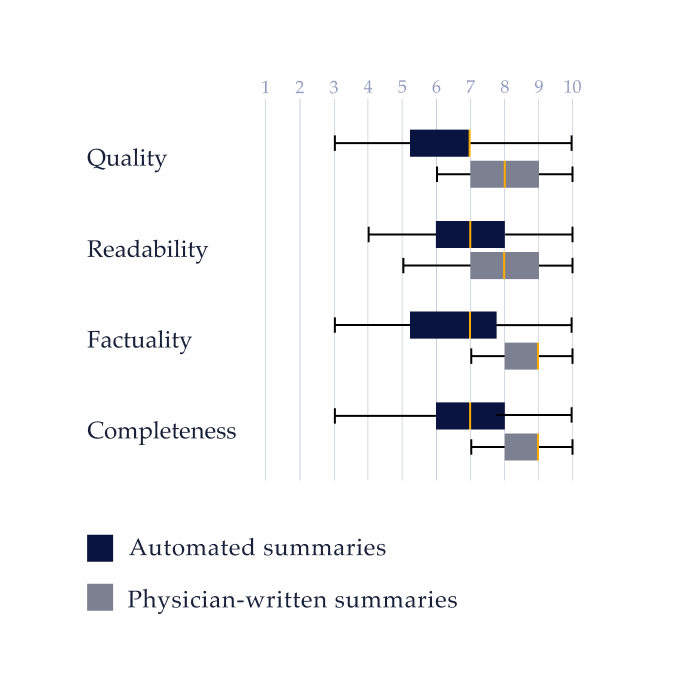}
    \label{fig:box_plot}
\end{figure}

As shown in Table \ref{tab:eval-results} and Figure \ref{fig:box_plot}, the quality of the automated summaries had an average rating of 6.52; 62\% met the quality standards of care, defined as a quality score of 7 or higher, compared to 94\% of the physician-written summaries. The automated summaries were also highly readable with a mean score of 7 out of 10; in fact, the two raters misclassified 34\% of the physician-written summaries as being generated by the automated method. The factuality (6.4 versus 8.6) and completeness (6.9 versus 8.7) ratings were slightly lower for the automated as compared with the physician-written summaries. 

\begin{table}[t!]
\small
\centering
\caption{\label{tab:eval-results}Mean and standard deviation (Sd.) of the ratings for quality, readability, factuality, and completeness of the automated and physician-written summaries. \\}
\begin{tabular}{p{35mm}x{35mm}x{35mm}}
\hline
 & \textbf{Automated} & \textbf{Physician-written} \\ \hline
Quality, \scriptsize Mean (+- Sd.) & 6.52 (+-1.50) & 8.16 (+-1.13) \\ \hline
Readability & 7.00 (+-1.46) & 8.00 (+-1.44) \\ \hline
Factuality & 6.44 (+-1.55) & 8.60 (+-0.88) \\ \hline
Completeness & 6.88 (+-1.70) & 8.68 (+-0.84) \\
\end{tabular}
\end{table}

Intraclass correlation coefficients were 0.64 for quality, 0.59 for readability, 0.73 for factuality, and 0.68 for completeness, which indicates moderate reliability of agreement for the two physician reviewers.

\section{Discussion}
\label{sec:discussion}
We found the automated approach did very well in mirroring the structure of a manually written physician summary (see example in Table \ref{tab:human-example} in the Appendix). Notably, 62\% of randomly sampled automated summaries met the standards of clinical validity, which suggests the approach is effective. Similar to how prior research has demonstrated clinical validity for summarizing the impression section for radiology reports \cite{zhang2019optimizing}, our results demonstrate how transformer models are effective at summarizing a patient’s hospital stay to generate the hospital course section of a discharge summary.

Note that in order to perform well on the quality metric, as our approach did, our technique had to perform well for every day of a patient’s stay in the hospital for conciseness, readability, factuality, and completeness and then select the most salient content within that time interval along with any clinical follow-ups for discharge. Therefore, the very high ROUGE scores for the HPI and daily segments and the high accuracy scores for the document classification and follow-up sentences all came together as an adequate measure for the quality performance of our automated summaries. Another contributing factor was the baseline performance of the TextRank method (a non-supervised approach) on the dataset; the high ROUGE scores on TextRank implied that our transformer models had high textual overlap, which has been previously shown to assist with factuality for abstractive summarization \cite{ladhak2021faithful}. 

While we demonstrated the success of our model in emulating a physician summary, we recognize several limitations and areas of future work, especially for  factuality and completeness. Of interest is that the physician clinical notes, such as Progress Notes and Consults, were overwhelmingly the primary source of content selected for summarization because of their high textual overlap with content in the discharge summary hospital course section; our model included very little source data from labs and vitals that were not contextualized within a physician note. Thus, if a physician did not include any context with respect to labs and vitals in their notes (such as documenting the implications of an abnormality for lab results), our model assumed that the data was more or less irrelevant. The implication is that our model, as constructed, was not adequate to find, interpret, and synthesize lab and vitals data that may have been missed by physicians in their clinical notes during the course of the patient’s hospital stay.

Given these limitations, we make a few recommendations. First, for improving the BERT-based document classification model (see Figure \ref{fig:data_flow}) which had an accuracy of 78.83\%, it should be paired with a rule-based structure to always summarize documentation for specific procedures, consults, and events (even if the findings are insignificant). For example, if a patient is administered a tissue plasminogen activator (tPA) drug,  commonly used for emergency treatment for ischemic stroke, the event should always be included in a hospital course summary. Our BERT-based document classification model approach could then be used for cases where it is too difficult to create such rules. 

Second, we recommend constructing a method for improving factuality for proper names, dates, numbers, and other similar items. As seen in the example in Table \ref{tab:human-example} in the Appendix, our approach occasionally misidentified the surgeon’s name who performed a procedure or swapped dates for when two procedures were performed. This occurred because our model hallucinated the most common surgeon’s name or dates. Since our approach was only designed to guarantee factuality of medical terminology, this kind of hallucination continues to be a known limitation.

Lastly, the two evaluators commented that the readability of the automated summaries could be improved by reducing the prevalence of acronyms. For background, physicians commonly write medical acronyms, including some that are not common outside their specialty, in their clinical notes. The acronyms are not always a one-to-one match with their respective expanded words. As our model was trained with physician-written discharge summaries as the gold-truth, these same acronyms continued forward in the automated summaries. To address this challenge, acronyms in the dataset could be manually converted to their respective expanded words. Alternatively, an additional trained model may be added that automatically translates medical acronyms to their correct expanded words based on the context in which the acronym was used.

Our study revealed several expanded applications of clinical note summarizations. While our dataset was specific to patients admitted to a neurology unit, the prior study that used a similar approach used the MIMIC-III dataset for ICU patients \cite{hartman2022discharge}. Thus, our day-to-day approach could potentially be adapted in the future to other inpatient clinical specialties if fine-tuned on a different dataset. Likewise our summary was constructed chronologically for the hospital stay by abstracting 1-2 sentences each day for the patient; so as new clinical content became available, the prior summarized sentences remained intact. The implication is that our approach could also be used to create a transfer report for patients moved from one medical unit to another before discharge. Finally, our study design demonstrated performance for complete automation of the discharge summary hospital course. In practice, a healthcare organization could add a step where physicians review a drafted automated summary and make slight corrections before finalization. Such a workflow would still be anticipated to provide significant benefit to physicians in mitigating physician burnout.

\section{Conclusion}
\label{sec:conclusion}
Transformers can perform state-of-the-art NLP tasks such as text summarization. We present an approach of using transformers, enhancing these models for clinical factuality by constraining medical terminology, and then dividing the medical chart into three separate segments to automate the hospital course section of the discharge summary. Through our work with an inpatient neurology EHR dataset, we have shown the potential of this approach as a means of constructing an automated patient summary of the hospital chart. Findings from this study could be used by a healthcare organization to determine the potential value of implementing clinical text summarization methods in a real-time production setting.

\section{Acknowledgements}
\label{sec:acknowledgements}

We are grateful for the guidance we received from Alexander “Sasha” Rush with respect to our approach of improving clinical factuality in a sequence-to-sequence transformer model. We are thankful for the assistance we received from Rita Giordana Pulpo for the graphical designs in our manuscript. We received support from NewYork-Presbyterian and Weill Cornell Medicine, including the Joint Clinical Trials Office and Clinical and Translational Science Center (UL1TR002384).

\bibliographystyle{vancouver}
\bibliography{literature}
\newpage

\appendix

\section{Criteria for Physician Evaluation}
\begin{table}[h!]
\begin{adjustwidth}{-1.5cm}{-1.5cm}
\small
\centering
\caption{\label{tab:assessment-criteria}Definitions for each metric provided to the two physicians for the clinical evaluation protocol.\\}
\begin{tabular}{|p{25mm}|p{125mm}|}
\hline
\textbf{Metric} & \textbf{Definition} \\ \hline
Quality & Rate the summary of what you view as a well-written discharge summary hospital course. Keep in mind that the intended purpose of the discharge summary, as set out by CMS and the Joint Commission, is to clearly communicate a patient's care plan to the post-hospital care team. \\ \hline
Readability & The language and grammar is what would be expected of a trained physician. The intended audience for the summary is the post-hospital care team (usage of medical terminology, abbreviations, and syntax is acceptable if a general physician could understand them). \\ \hline
Factuality & All information presented is accurate and factually correct. \\ \hline
Completeness & The summary includes the relevant clinical events that occurred during the patient stay. Assess the summary on the standard for not missing any key clinical information that would be detrimental to the patient's post hospital care if the downstream provider were not informed. \\ \hline
\end{tabular}
\end{adjustwidth}
\end{table}

\pagebreak

\section{Example Discharge Summary Hospital Courses}
\begin{table}[h!]
\footnotesize
\centering
\caption{\label{tab:human-example}Example of an automated summary and a separate physician-written summary of the hospital chart that were evaluated by two board-certified physicians with the criteria listed in Table \ref{tab:human-evaluation}. PHI has been redacted.\\}
\begin{tabular}{|C{16mm}|C{16mm}|C{16mm}|C{16mm}|C{16mm}|C{16mm}|C{16mm}|C{16mm}|}
\hline
\multicolumn{4}{|c|}{\textbf{Automated Summary}} & \multicolumn{4}{c|}{\textbf{Physician-Written Summary}} \\ \hline
\multicolumn{4}{|p{75mm}|}{[AGE] year old woman with history of HTN, HLD, Hypothyroidism, and hyperparathyroidism who presented to GBG Adult Emergency on [DATE] after having an incidental findings of an expansile, peripherally hyperdense, centrally hypodense mass in the left paramedian frontal lobe and extending contralaterally across the midline. While in the emergency department the patient was seen by neurosurgery. MRI brain without contrast was completed on [DATE] which showed a large, centrally necrotic, extra-axial mass centered along the right mid falx with broad-based dural attachment of the left anterior falx and the left frontal convexity, measuring 6.1 x 4.2 cm (AP x TV x CC; series 800, image 51). Ophthalmology was consulted on [DATE] to rule out disc edema. Taken to OR for Left frontal craniotomy for resection of parasagittal meningioma (a modifier 22 will be added given the massive size of the lesion which was encased in pericallosal artery) on [DATE] with Dr. [PHYSICIAN]. Pt deemed stable for transfer to Neurosurgery floor [DATE]. Patient deemed safe for discharge home on [DATE]. }  & \multicolumn{4}{p{75mm}|}{[AGE] year old female with headaches and multiple punctate subacute strokes, and presumed cerebral amyloid angiopathy. She presented on [DATE] with confusion for 2 days as well as lethargy. CTH founds a R frontal ICH extending into corpus callosum with 2mm midline shift. Platlets 333. She was admitted to the Neuro-ICU for close monitoring. She was treated with nicaridpine ggt for BP goal. She was seen by neurosurgery who did a brain biopsy of her right temporal area [DATE] to differentiate inflammatory CAA disease vs non-inflammatory phenotypes. She was then treated empirically with IV solumedrol 1g daily [DATE]. Her neurological symptoms improved and she was transferred to stepdown neurology. She requested to have her final doses at home as an outpatient and was discharged home to receive her final doses of methylprednisone at home on [DATE] with an infusion company. She will follow with Dr [PHYSICIAN] on discharge. Home infusion for completion of Solumedrol x 2 days.Home services to include: nursing for assessment and teaching,PT for gait training and OT for ADL training.}  \\ \hline
\multicolumn{8}{|c|}{\textbf{Average Evaluation Ratings}} \\ \hline
\textbf{\scriptsize Quality} & \textbf{\scriptsize Readable} & \textbf{\scriptsize Factual} & \textbf{\scriptsize Complete} & \textbf{\scriptsize Quality} & \textbf{\scriptsize Readable} & \textbf{\scriptsize Factual} & \textbf{\scriptsize Complete} \\ \hline
7.5 & 8.5 & 8 & 8 & 8.5 & 9 & 9.5 & 8 \\ \hline
\multicolumn{8}{|c|}{\textbf{Physicians' Review Comments}} \\ \hline
\multicolumn{4}{|p{75mm}|}{The surgery was performed by a different physician. The ophthalmology findings are missing. An incidental finding from brain imaging should reference the reason she was having brain imaging in the first place.}  & \multicolumn{4}{p{75mm}|}{Echo results missing. MRI brain, MRA head, CTA head results missing. In mentioning the need for a nicardipine drip, it would have helped see information about the blood pressures during the admission that may have required that intervention.}  \\ \hline
\end{tabular}
\end{table}

\end{document}